\documentclass[final]{l4dc2026}

\title[MATT--Diff]{MATT--Diff: Multimodal Active Target Tracking by Diffusion Policy}
\usepackage{times}
\usepackage{algorithm}
\usepackage{algorithmic}

\newcommand{\bfa}{\mathbf{a}}

\newcommand{\bfo}{\mathbf{o}}
\newcommand{\bfp}{\mathbf{p}}

\newcommand{\bfu}{\mathbf{u}}

\newcommand{\bfw}{\mathbf{w}}
\newcommand{\bfx}{\mathbf{x}}
\newcommand{\bfy}{\mathbf{y}}

\newcommand{\bfmu}{\boldsymbol{\mu}}

\newcommand{\bbR}{\mathbb{R}}

\newcommand{\calI}{\mathcal{I}}

\newcommand{\calN}{\mathcal{N}}

\author{%
 \Name{Saida Liu$^1$} \Email{2215072t@stu.kobe-u.ac.jp}
 \AND
 \Name{Nikolay Atanasov$^2$} \Email{natanasov@ucsd.edu}
 \AND
 \Name{Shumon Koga$^1$}
\Email{koga@harbor.kobe-u.ac.jp}\\
\addr $^1$Department of Computer Science and Systems Engineering, Kobe University, Kobe, Hyogo, 657-8501, Japan \\
\addr $^2$Department of Electrical and Computer Engineering, University of California San Diego, La Jolla, CA, 92093, USA%
}

\begin{document}

\maketitle

\begingroup
  \makeatletter
  \renewcommand\thefootnote{}
  \renewcommand\@makefntext[1]{#1}
  \footnotetext{
  This work was supported by JST-Research \& Development Program for
Next Generation Edge AI Semiconductors Japan Grant Number JPMJES2514
}%
  \addtocounter{footnote}{-1}
  \makeatother
\endgroup

\begin{abstract}
This paper proposes MATT-Diff: Multimodal Active Target Tracking by Diffusion Policy, a control policy for active multi-target tracking using a mobile agent. The policy enables multiple behavior modes for the agent, including exploration, tracking, and target reacquisition, without prior knowledge of the target numbers, states, or dynamics. Effective target tracking demands balancing exploration for undetected or lost targets with exploitation, i.e., uncertainty reduction, of detected but uncertain ones. We generate a demonstration dataset from three expert planners including frontier-based exploration, an uncertainty-based hybrid planner switching between frontier-based exploration and RRT* tracking, and a time-based hybrid planner switching between exploration and target reacquisition based on target detection time.
Our control policy utilizes a vision transformer for egocentric map tokenization and an attention mechanism to integrate variable target estimates represented by Gaussian densities. Trained as a diffusion model, the policy learns to generate multimodal action sequences through a denoising process. 
Evaluations demonstrate MATT-Diff's superior tracking performance against other learning-based baselines in novel environments, as well as its multimodal behavior sourced from the multiple expert planners. Our implementation is available at \url{https://github.com/CINAPSLab/MATT-Diff}. 
\end{abstract}


\section{INTRODUCTION} \label{sec:introduction}
The ability to track targets actively using mobile robots has widespread applications in security and surveillance, environmental monitoring, and search and rescue \citep{queralta2020collaborative}.
The complexity of this task stems from several interconnected challenges: (i) effectively exploring the environment to detect and acquire targets; (ii) accurately predicting target motion to sustain tracking under noisy observations and unmodeled dynamics; (iii) planning 
agent trajectories that actively minimize target uncertainty under limited field of view (FoV) constraints; and (iv) adaptively modifying target beliefs and planning strategies when targets are not detected where they are expected to be.
Optimization- and sampling-based planners yield single, deterministic solutions, struggling with the multi-modal decisions inherent in uncertain real-world scenarios, particularly the exploration-exploitation dilemma. 
Learning policies that handle multi-modal action distributions is crucial in the contex of target tracking.

This paper proposes MATT-Diff: Multi-Modal Active Target Tracking by Diffusion Policy. Our approach addresses 
agent control in complex target tracking scenarios without prior knowledge of target numbers, states, or dynamics in known environments with obstacles and limited FoV. The target states are estimated using Kalman filters with updates only when targets are within the FoV. We generate a diverse dataset of demonstrations from three planners: a frontier-based exploration searching for undetected targets and two hybrid planners combining exploration with RRT* tracking of uncertain targets based on uncertainty or detection time.
Our MATT-Diff control policy employs a vision transformer network, processing egocentric occupancy-grid maps via tokenization and an attention mechanism robustly integrating a variable number of target estimates.
The policy is trained using diffusion to match the multi-modal action sequences from expert demonstrations through a denoising process, embedding the processed tokens. 
Through numerical experiments, we demonstrate MATT-Diff's superior performance to other learning-based baselines across randomized numbers and motions of targets in an environment map not included in training data. 

\section{RELATED WORK} \label{sec:relatedwork}
We review related work on target tracking and diffusion models.

\subsection{Active Target Tracking}

Active target tracking has a rich history in robotics, originating from early studies on pursuit-evasion games \citep{lavalle1997motion} and later evolving into the area of sensor planning and management \citep{spletzer2003dynamic}. To handle sensor noise and uncertainty in a probabilistic formulation, information-theoretic approaches were introduced, e.g.
in \citet{le2009trajectory,hero2011sensor} proposing nonmyopic trajectory planning for targets with known linear Gaussian dynamics. This approach has since been extended to multi-robot systems, addressing challenges such as resilience, anytime planning, asymptotic optimality, and communication constraints \citep{zhou2018resilient, schlotfeldt2018anytime, kantaros2019asymptotically, wang2024uncertainty}. An informative path planning framework has been proposed in \cite{meera2019obstacle} for target search and in \cite{sudha2025informative} for dynamic occupancy mapping. Low-level control for handling occlusion in maintaining target's visibility has been proposed in \citet{zhou2025control}. For a comprehensive overview of target tracking, particularly in the context of unmanned aerial vehicles, see \citet{sun2024moving}.

More recently, the field has shifted towards learning-based methods that leverage deep neural networks. Deep reinforcement learning (RL) has been employed to train policies for active target tracking, enabling robots to learn complex behaviors directly from experience or simulation by model-free \citep{jeong2021deep}, model-based \citep{Yang_MBRL_ActivePerception_L4DC23}, and graph-neural network (GNN) approaches \citep{10160723}. While these methods demonstrate effective learning capabilities, they often produce unimodal policies that can struggle in scenarios requiring diverse, context-dependent strategies. Our work builds on learning-based methods by specifically addressing the need for multi-modal action generation, a gap not fully explored by existing RL approaches.

\subsection{Diffusion Models}

Diffusion models have recently emerged as a powerful representaiton of multi-modal action distributions in robotics \citep{wang2023diffusion}, following their success in generative AI \citep{song2021scorebased}. 
Behavior cloning (BC; \citealp{ijcai2018p687}) relies on supervised imitation of expert demonstrations, but its unimodal policy representation often collapses multi-modal action distributions into averaged behaviors, leading to degraded performance even without covariate shift \citep{zare2024survey}.

Through iterative denoising, diffusion models learn to generate coherent action sequences conditioned on past observations, enabling policies to reproduce diverse expert strategies.
Diffusion policies have demonstrated remarkable success in high-dimensional control tasks, including visuomotor control \citep{chi2023diffusion}, language-conditioned multi-task policies \citep{yan2024dnact}, and dexterous robot manipulation \citep{song2025survey}. 
Recent work has also incorporated geometric symmetries such as rotation into diffusion policies to improve generalization and sample efficiency in tasks with spatial invariances \citep{wang2024equivariant}.
Integrating diffusion models with reinforcement learning has been explored both in online training \citep{ding2024diffusionbased} and in fine-tuning of pre-trained models \citep{ren2025diffusion, li2024learning, wagenmaker2025steering}.

The application of diffusion policies to mobile robot navigation is a burgeoning area of research. A pioneering work in this domain is NoMaD \citep{sridhar2024nomad}, which proposes a unified policy for both goal-directed navigation and goal-agnostic exploration using a goal-masking mechanism within a transformer architecture. 
In DARE \citep{cao2025dare}, a diffusion policy for exploration was trained to reason about partially observed environments—i.e., to act based on incomplete belief states—
using expert demonstrations generated with access to the full ground-truth map.
Other works have applied diffusion models not for direct policy learning but for auxiliary tasks in visual navigation, such as generating goal proposals \citep{shah2023vint} or trajectories \citep{zeng2025navidiffusor}. Our work, MATT-Diff, contributes to this line of research by formulating and training a diffusion policy for active multi-target tracking, where the policy must learn to balance exploration for new/lost targets with exploitation of currently tracked targets in a multi-modal fashion.

\section{PROBLEM STATEMENT} \label{sec:problem}
Consider a mobile agent with state $\mathbf{x}_t \in \mathbb{R}^{n_x}$ and control input 
$\mathbf{u}_t \in \mathbb{R}^{n_u}$, evolving in discrete time according to $\mathbf{x}_{t+1} = f(\mathbf{x}_t, \mathbf{u}_t)$. 
The agent operates in an environment containing $N_y$ targets with individual states $\mathbf{y}_t^{(j)} \in \mathbb{R}^{n_y}$ for $j \in \{1, \dots, N_y\}$. Both the exact number of targets $N_y$ and their dynamics are unknown to the agent. The agent is equipped with an onboard sensor with a limited FoV denoted by $\mathcal{F}(\mathbf{x}_t) \subset \bbR^3$ which is dependent on the agent state due to occlusion. 
The sensor provides measurements $\mathbf{z}_t^{(j)}$ of the $j$-th target if and only if its position $\mathbf{p}(\mathbf{y}_t^{(j)}) \in \mathbb{R}^3$ lies within the agent's FoV at time $t$. The sensor model is given by
\begin{equation}
    \mathbf{z}_t^{(j)} = H \mathbf{y}_t^{(j)} + \boldsymbol{\eta}_t, 
    \quad \text{if } \quad \mathbf{p}(\mathbf{y}_t^{(j)}) \in \mathcal{F}(\mathbf{x}_t),
\end{equation}
where $\boldsymbol{\eta}_t \sim \mathcal{N}(0, R)$ is Gaussian measurement noise with covariance $R \in \mathbb{R}^{n_z \times n_z}$.

To estimate the target states, the agent employs a Kalman filter with the limited FoV constraint. The state of each hypothesized target $j$ is estimated as a Gaussian $\mathbf{y}_t^{(j)} | \mathbf{z}_{0:t} \sim \mathcal{N}(\bfmu_t^{(j)}, \Sigma_t^{(j)})$. The filter update step is executed only when the corresponding target is successfully detected and measured within the agent's FoV. Regardless of the target detection, the filter continues with its prediction step, where the matrices $(A, W)$ describing the target dynamics $\bfy^{(j)}_{t+1} = A \bfy^{(j)}_t + \bfw^{(j)}_t$ with $\bfw^{(j)}_t \sim \calN(\mathbf{0}, W)$ are hypothesized. 
Due to the process noise $\bfw^{(j)}_t$, if a target is not detected, its covariance grows to $\Sigma_{t+1}^{(j)} = A\Sigma_t^{(j)}A^\top + W$ and its determinant reflects the increased uncertainty.

A critical challenge arises in the scenario of missing targets. This occurs when the agent predicts a target to be within its FoV, i.e., $\mathbf{p}(\bfmu_t^{(j)}) \in \mathcal{F}(\mathbf{x}_t)$, but the true target state $\mathbf{y}_t^{(j)}$ is not actually within the FoV, i.e., $\mathbf{p}(\mathbf{y}_t^{(j)}) \notin \mathcal{F}(\mathbf{x}_t)$.
In this case, the agent temporarily ignores the target, 
while its Kalman filter continues prediction updates and resumes measurement updates upon re-detection without reinitialization.
Overall, the set of detected targets $\calI^{(D)}_t$ up to time $t$ is given by $\calI^{(D)}_t = \calI^{(D)}_{t-1} \cup \calI^+_t \setminus \calI^-_t$, where $\calI^+_t := \{ j \in \{1, \dots, N_y\} | \mathbf{p}(\mathbf{y}_t^{(j)}) \in \mathcal{F}(\mathbf{x}_t)\}$ denotes the set of newly discovered targets at time $t$, and $\mathcal{I}^-_t := \left\{ j \in \mathcal{I}^{(D)}_t \setminus \mathcal{I}^+_t \hspace{1mm} \bigg| \hspace{1mm}\mathbf{p}(\mu_t^{(j)}) \in \mathcal{F}(\mathbf{x}_t) \right\}$ denotes the set of lost targets that were previously detected but are now missing from the FoV at time $t$.

The agent faces an \emph{exploration-exploitation dilemma}: should it continue to ''exploit'' by moving towards detected targets to minimize the uncertainty in their states or should it ``explore'' the environment to search for lost targets or discover new previously unobserved targets? 
This paper addresses the problem of learning to autonomously navigate and make these multimodal decisions to achieve effective target tracking.

\section{METHODOLOGY} \label{sec:methods}
Our methodology is centered around learning a multimodal diffusion policy from demonstrations collected by expert target-tracking methods. We first describe the design of the expert algorithms, which generate rich demonstration data. We, then, detail the architecture of our control policy and the diffusion-based training process.

\subsection{Expert Planners for Multimodal Demonstrations}\label{sec:expert}

We describe three planners that can be used to generate a diverse dataset of target-tracking behaviors, which demonstrate the trade-off between exploration and exploitation effectively.

\paragraph{Frontier-Based Exploration Planner}

For a pure exploration planner, we utilize a well-known frontier-based occupancy map exploration approach \citep{yamauchi1997frontier}. We represent the environment as a discrete occupancy grid $\mathcal{E}\subset\mathbb{Z}^2$ with probabilistic occupancy values $\mathbf{M}_{\text{prob}}(i,j)\in[0,1]$, 
where a cell $(i,j)$ is classified as free if $\mathbf{M}_{\text{prob}}(i,j)<0.5$, unknown if $\mathbf{M}_{\text{prob}}(i,j)=0.5$, and occupied if $\mathbf{M}_{\text{prob}}(i,j)>0.5$. 
The obstacle and free regions are defined as $\mathcal{O}=\{\mathbf{p}\in\mathcal{E}\mid \mathbf{M}_{\text{prob}}(\mathbf{p})>0.5\}$ and $\mathcal{E}_{\text{free}}=\mathcal{E}\setminus\mathcal{O}$, respectively. 
Given our assumption that the ground-truth map is known, the free space $\mathcal{E}_{\text{free}}$ is divided into ''explored'' $\mathcal{E}_{\text{explore},t}$ 
and ''unexplored'' $\mathcal{E}_{\text{free}}\setminus\mathcal{E}_{\text{explore},t}$ regions. 
The frontier region is defined as $\mathcal{E}_{\text{frontier},t}=\partial\mathcal{E}_{\text{explore},t}\setminus\mathcal{O}$, representing the interface between known free and unknown space. The frontier exploration planner selects the next frontier by score minimization
$\mathbf{p}_t^* \in \arg\min_{\mathbf{p}\in\mathcal{E}_{\text{frontier},t}} S_t(\mathbf{p})$,
where $S_t(\mathbf{p})$ is a weighted combination of distance, visitation frequency, and expected coverage gain. A collision-free path to $\mathbf{p}_t^*$ is planned with RRT* \citep{karaman2011sampling} within a safety margin from obstacles and is tracked using curvature-based lookahead control. 
At each step, frontier points are scored based on a weighted function of distance, visitation frequency, and expected coverage gain, encouraging persistent exploration before the entire map is covered. Once the environment becomes fully explored, low-penalty revisits are favored instead of stalling, allowing the agent to reobserve previously lost or moving targets.

\paragraph{Uncertainty-Based Hybrid Planner}
Once at least one target is detected, a planner must decide whether to continue exploring for new targets or to track existing ones. We design a hybrid exploration-exploitation planner that makes this decision based on the targets' uncertainty. Namely, when all currently detected target states are known with high confidence (i.e., low uncertainty), the planner prioritizes exploration to discover new targets. Conversely, if any target's uncertainty exceeds a predefined threshold, the planner switches to tracking mode, prioritizing the reduction of uncertainty for the most unconfidently localized target. 
The target uncertainty is measured by its differential entropy, which is proportional to the log-determinant of its covariance matrix, $\log \det (\Sigma_t^{(j)})$ \citep{le2009trajectory}. In the tracking mode, the planner uses RRT* to generate a path toward the mean of the most uncertain target 
$\bfmu_t^{(j^*)}$, where $j^* = \arg\max_{j \in \mathcal{I}^{(D)}_t} \log \det (\Sigma_t^{(j)})$.

\paragraph{Time-Based Hybrid Planner}
As a third expert, we use a time-based hybrid exploration-exploitation planner. The planner first starts with frontier-based exploration and, once a target is detected, to estimate its state as accurately as possible, the planner keeps tracking the detected target for a fixed-time interval. Tracking is done via RRT* with goal iteratively set to mean of the tracked target. After the fixed time has passed, the planner switches back to frontier-based exploration to search for undiscovered targets.

\begin{figure*}[t]
\centering
\includegraphics[width=0.99\linewidth]{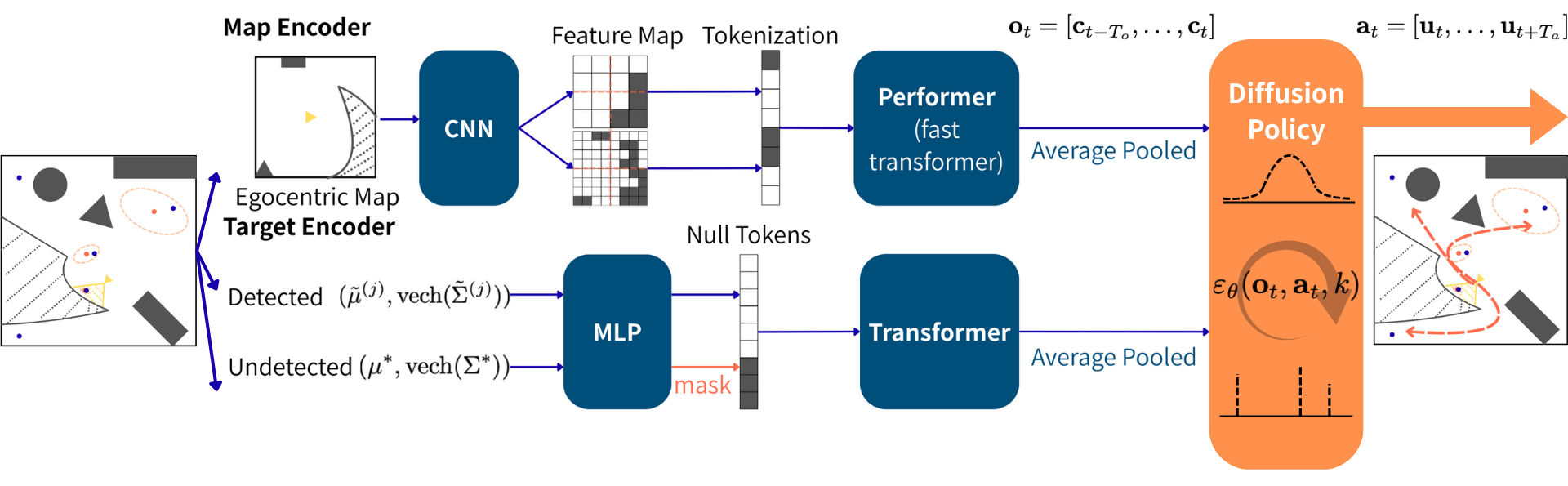}
\caption{Our MATT-Diff architecture consists of a map encoder and a target encoder. The map encoder converts a local egocentric map into patch tokens via CNN and feeds them into a Performer transformer. The target encoder processes detected target beliefs with masking for undetected targets through self-attention to produce context-aware embeddings. A diﬀusion policy performs a denoising process to generate multimodal action sequences.}\label{fig:network}
\end{figure*}

\subsection{Policy Network Architecture}\label{subsec:network}

The MATT-Diﬀ policy network is designed to process heterogeneous sensor and target-density inputs to generate a sequence of future agent actions. The agent pose is used only for coordinate transformations within the encoders to obtain egocentric representations. The architecture consists of two encoders, a map encoder and a target encoder, whose outputs are fused and used to condition a U-Net \citep{ronneberger2015u} in the diﬀusion model to produce multi-modal action sequences through iterative denoising. An overview of the architecture is shown in Fig.~\ref{fig:network}.

\paragraph{Map Encoder}
An egocentric occupancy-grid map, representing a subset of the global map of the environment in the agent’s local coordinate frame, is generated from the ground-truth occupancy-grid map and the current agent pose.
Specifically, the global map is transformed by the agent pose $\mathbf{x}_t = [p_t, \theta_t]$ through rotation and translation to obtain an egocentric view centered at the agent’s position $p_t$ and aligned with its orientation $\theta_t$.
This local map is then processed by a Convolutional Neural Network (CNN) to extract multi-resolution features.
The features are divided into a sequence of non-overlapping patches, i.e., map tokens, which are linearly embedded and passed to a transformer encoder to obtain a latent representation of the local environment geometry \citep{pmlr-v205-xiao23a}. We use a Performer \citep{choromanski2021rethinking}, a fast transformer model with linear time and space complexity, to efficiently handle the high dimensionality of the egocentric map.

\paragraph{Target Encoder}
To handle a variable number of targets, we use an attention module. For a detected target's Gaussian density $(\boldsymbol{\mu}, \Sigma)$, the encoder uses an input feature $(\tilde{\boldsymbol{\mu}}, \tilde{\Sigma}) = (\mathbf{q}(\mathbf{x}, \boldsymbol{\mu}), \frac{\Sigma}{\log \det( \overline{\Sigma})} )$, where $\mathbf{q}(\mathbf{x}, \mathbf{p})$ transforms the position $\mathbf{p}$ into the agent's coordinate frame given the agent pose $\mathbf{x}$, and $\overline{\Sigma}$ is a threshold covariance matrix defined later. Conversely, for undetected targets, the encoder input is set as $(\boldsymbol{\mu}^*, \Sigma^*) = (\mathbf{q}(\mathbf{x}, \mathbf{0}), \frac{\overline{\Sigma}}{\log \det( \overline{\Sigma})} )$. This representation centers the target belief at the environment's origin (viewed from the agent's frame) and assigns a sufficiently large covariance $\overline{\Sigma}$ so that the target belief is uninformative by covering nearly the entire environment. The state vectors for both detected and undetected targets are passed through a Multi-Layer Perceptron (MLP) to create target embeddings, which are then processed by a self-attention layer (transformer encoder). During this step, undetected targets are masked based on the condition $\log \det (\Sigma^*) \geq 1$. Finally, mean pooling is applied to these representations to generate a single fixed-size context vector that summarizes the entire multi-target set.

\subsection{Diffusion Policy Training}

Our policy is trained as a conditional Denoising Diffusion Probabilistic Model (DDPM) \citep{ho2020denoising} to generate a $T_a$-step future action sequence $\mathbf{a}_t = [\mathbf{u}_t, \dots, \mathbf{u}_{t+T_a -1}]$ given a $T_o$-step past observation context sequence $\bfo_t = [\mathbf{c}_{t- T_o + 1},  \dots, \mathbf{c}_{t}]$, where each $\mathbf{c}_t$ denotes the fused map-target embedding obtained from the encoders described in Section~\ref{subsec:network}. The network $\boldsymbol{\varepsilon}_{\theta}$ is trained to predict the added Gaussian noise 
at each diffusion timestep $k$. 
Here, $\boldsymbol{\varepsilon}^k$ denotes the sampled noise at step $k$, 
and $\bfa_t^k = \bfa_t + \boldsymbol{\varepsilon}^k$ represents the noisy version of the clean expert action sequence $\bfa_t$. 
The training objective minimizes the mean squared error between the true and predicted noise:
\begin{align}
    \mathcal{L}(\theta) = 
    \| \boldsymbol{\varepsilon}^k - \boldsymbol{\varepsilon}_{\theta}(\bfo_t, \bfa_t^k, k) \|^2.
\end{align}
At inference time, an action sequence is generated by starting with pure noise $\mathbf{a}_K \sim \mathcal{N}(0, \mathbf{I})$ and iteratively applying the learned denoising function for $k = K, \dots, 1$ by 
\begin{align}
    \bfa_{t}^{k-1} = \alpha (\bfa_t^{k} - \gamma \varepsilon_{\theta} (\bfo_t, \bfa_t^k, k) + \calN(0, \sigma^2 I) ),
\end{align}
where $\alpha, \gamma, \sigma>0$ are noise scheduling functions of $k$. This iteration produces a clean action plan $\mathbf{a}_t^0$ which is implemented as an output of the diffusion policy.
\section{EVALUATION} \label{sec:evaluations}

We train MATT-Diff and compare its performance against a behavior cloning (BC) baseline~\citep{ijcai2018p687}, a deep reinforcement learning (RL) baseline adapted from~\citet{jeong2021deep}, and an ablation model of MATT-Diff eliminating the map encoder to justify the contribution of the egocentric location awareness provided in our architecture.
The BC baseline learns a deterministic policy $\pi_{\phi}$ that predicts actions $\mathbf{u}_t$ from observations $\mathbf{o}_t$ (encoded map and target beliefs). The parameters $\phi$ are optimized to minimize the mean squared error
$\mathcal{L}_{\mathrm{BC}} = \left\| \mathbf{u}_t^{\text{exp}} - \pi_{\phi}(\mathbf{o}_t) \right\|^2$, where $\mathbf{u}_t^{\text{exp}}$ denotes expert actions. The RL basline uses the Deep Q-Network (DQN) for target tracking proposed by \citet{jeong2021deep}. The agent is trained by setting the reward to maximize as $ R_t = - \lambda \sum_{j \in \mathcal{I}_t^{(D)}} \log(\det(\Sigma_{t}^{(j)})$, namely, minimizing the differential entropy, where $\lambda >0$ is a scaling factor to stabilize the DQN learning process (set as $\lambda = 0.1$ in our experiments). MATT-Diff (w/o map encoder) is implemented based solely on the target encoder to generate actions, serving to validate the need for map tokenization. The training, simulation, and evaluation are done on a computer with Ubuntu 24.04, Intel Core Ultra 7, Nvidia GeForce RTX 5080.  

\subsection{Experiment Setup}

The experiment setup for both training and evaluation is described below.

\paragraph{Agent dynamics} The agent's dynamics are set to a two-dimensional single-integrator model, i.e., the agent's position $\bfp_t \in \bbR^2$ follows $\bfp_{t+1} = \bfp_t + \bfu_t$ with control input $\bfu_t \in \bbR^2$, in both training and evaluation scenarios. To represent FoV, the heading angle is set to $\theta_{t+1} = \arctan (u_t^{(2)} / u_t^{(1)}) $ where $u^{(i)}_t \in \bbR$ for $i \in \{1,2\}$ is $i$-th element in $\bfu_t$. 
\paragraph{Target configulations and estimation}
The target dynamics are set to follow a Brownian velocity model in 2-D, i.e., with target dimension $n_y = 2$, transition matrix $A = I$, and process noise covariance $W = \operatorname{diag}(w_x^2, w_y^2)$, where $(w_x, w_y)$ are sampled uniformly from $[0.8, 1.2]$ at the beginning of each episode. The sensor model is set as $H = I$ and the measurement noise covariance is set as $R = \operatorname{diag}(r_x^2, r_y^2)$ with $r_x = r_y = 0.05$.
We consider the case where the ground-truth noise covariances in targets' process noise and the measurement noise are unknown to the agent. 
To represent a conservative state estimation in the agent's internal Kalman filter, 
the estimated process noise covariance $
\hat{W}$ is set to $\hat{W} = \operatorname{diag}(90, 40)$ and the estimated measurement noise covariance $\hat{R}$ is set to a significantly higher magnitude than the ground-truth. For each configuration, we run 20 randomized episodes across the unseen map and report the mean and the standard deviation of the performance metrics. Both the training and evaluation scenarios follow this setup across randomized number of moving targets among $N_y \in \{3, \dots, 6\}$ and randomized initial configurations for both agent's and targets' states. 

\paragraph{Environment maps} The environment maps are sourced from the HouseExpo dataset \citep{li2020houseexpo}. The expert data generation by the three planners introduced in Section \ref{sec:expert} and the training of MATT-Diff, MATT-Diff (w/o map encoder), BC, and DQN are done over four different maps, while the evaluation is done on a map not included in the training maps to evaluate the performance in an out-of-distribution (OOD) environment.

\subsection{Episode-Averaged Results}

The performance of MATT-Diff, MATT-Diff(w/o map encoder), DQN, and BC is measured using three metrics: root mean squared error (RMSE), negative log-likelihood (NLL) \citep{pinto2021uncertainty}, and differential entropy of the target estimates. For undetected targets, as handled by the target encoder introduced in Section \ref{subsec:network}, the Gaussian densities are set to $\calN(\mathbf{0}, \overline{\Sigma})$ with sufficiently large $\overline{\Sigma}$, to approximate a uniform distribution over the environment. For detected targets with densities $\calN(\bfmu^{(j)}, \Sigma^{(j)})$ for $j \in \calI^{(D)}$ and ground-truth target states $\{\bfy^{(j)}\}$ for $j \in \calI = \{ 1, \dots, N_y\}$, the evaluation metrics are given by 
  $\textrm{RMSE} =  \sum_{j \in \calI^{(D)}} ||\bfy^{(j)} - \bfmu^{(j)} || / | I^{(D)} |$, $ 
    \textrm{Entropy} = \sum_{j \in \calI^{(D)}}  \log \left( \det \left( \Sigma^{(j)} \right) \right) 
    + (N_y - | \calI^{(D)} |) \log \left( \det \left( \overline{\Sigma} \right) \right) 
    $, and $$ 
    \textrm{NLL} =   - \sum_{j \in \calI^{(D)}} \log \left(  p_{\cal N}\left(\bfy^{(j)} | \bfmu^{(j)}, \Sigma^{(j)} \right) \right) 
    - \sum_{j \in \calI \setminus \calI^{(D)}} \log \left(  p_{\cal N}(\bfy^{(j)} | \mathbf{0}, \overline{\Sigma} ) \right),$$ 
%
where $p_{\cal N}(\bfx | \bfmu, \Sigma)$ is the probability density function of a Gaussian with mean $\bfmu$ and covariance $\Sigma$. For all three metrics, lower values are desirable for successful target tracking. 

We evaluate the trained MATT-Diff, MATT-Diff (w/o map encoder), DQN, and BC by analyzing the average and standard deviation of the metrics over 20 randomized episodes in an OOD map, summarized in Table~\ref{tab:map2}.
As shown in Table~\ref{tab:map2}, MATT-Diff achieves the lowest averaged value in all metrics among all the implemented learning-based approaches, exhibiting the strongest zero-shot generalization to an OOD environment. 
This observation justifies the MATT-Diff's capability to effectively fuse spatial awareness and target densities to balance exploration and tracking in a multimodal fashion.

\begin{table}[t]
    \centering
    \caption{The average and standard deviation of the performance metrics over 20 randomized episodes in an out-of-distribution (OOD) map. The lowset average values (i.e., the best results) are highlighted in bold. MATT-Diff outperforms all baselines across all metrics.}
    \label{tab:map2}
    \begin{tabular}{lccc}
        \hline
        Method & RMSE & NLL & Entropy \\
        \hline
        \textbf{MATT-Diff} 
            & \textbf{268.840 $\pm$ 155.540} 
            & \textbf{13.124 $\pm$ 1.468} 
            & \textbf{13.332 $\pm$ 1.017} \\
        
        MATT-Diff (w/o map encoder)
            & 273.525 $\pm$ 148.701 
            & 13.295 $\pm$ 1.321 
            & 13.501 $\pm$ 0.911 \\
            
        BC
            & 299.776 $\pm$ 133.638 
            & 13.464 $\pm$ 1.351 
            & 13.582 $\pm$ 1.014 \\
            
        DQN
            & 297.049 $\pm$ 134.074 
            & 13.335 $\pm$ 1.878 
            & 13.465 $\pm$ 1.589 \\
        \hline
    \end{tabular}
\end{table}

\subsection{Per-Episode Results}\label{sec:Per-Episode Results}

\begin{figure}[t]
    \centering
    \includegraphics[width=0.99\linewidth]{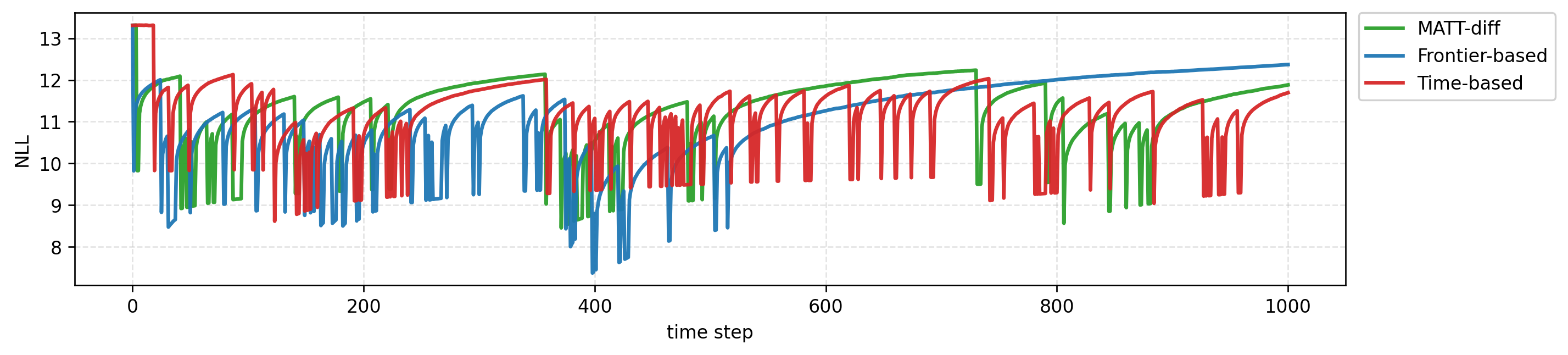}
    \caption{Temporal evolution of NLL for the frontier-based (blue) expert, the time-based (red) expert, and MATT-diff (green) in one episode.}\label{fig:time-metrics}
\end{figure}

\begin{figure*}[t]
  \centering
  \includegraphics[
    width=0.99\linewidth
  ]{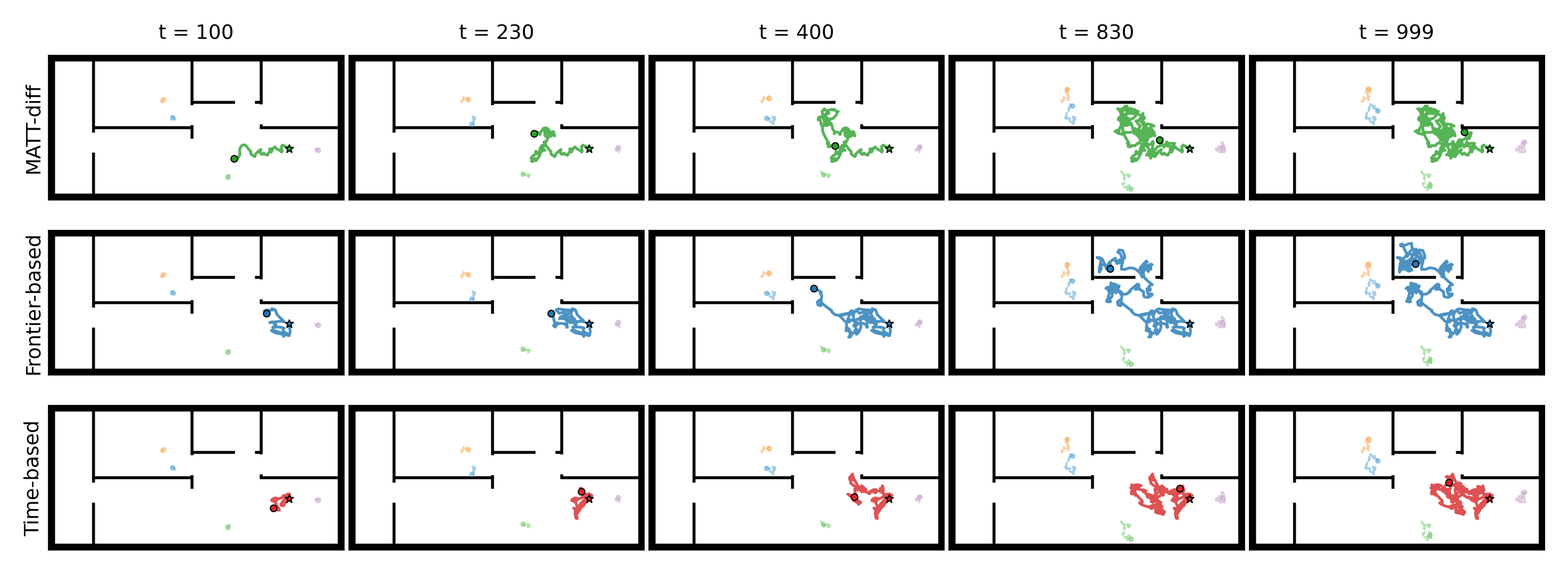}
  \caption{Trajectory snapshots of MATT-Diff and the expert planners over a single episode. MATT-Diff achieves a good balance of exploration, target tracking, and re-acquisition.}
  \label{fig:trajectories_map2}
\end{figure*}

We analyze the temporal behavior and performance metrics for MATT-Diff and the expert planners.
Fig.~\ref{fig:time-metrics} and Fig.~\ref{fig:trajectories_map2} show the temporal evolution of NLL and the corresponding trajectory visualizations for MATT-Diff, the frontier-based expert, and the time-based expert within a single episode. Although this specific episode does not represent the peak performance, it was selected for its clear illustration of the distinct behavioral phases inherent in our MATT-Diff policy. Specifically, the NLL curve of MATT-Diff (green) in Fig.~\ref{fig:time-metrics} exhibits six characteristic phases: (1) an initial \emph{exploration} phase, where after an early detection, the NLL rises as the agent prioritizes exploring the frontier space rather than immediate tracking, showing a trend similar to the frontier-based planner (timesteps 0--75); (2) a \emph{tracking} phase, where the NLL drops sharply upon target traking and remains low (timesteps 75--100); (3) a renewed \emph{exploration} phase, marked by a steady, monotonic increase in NLL as the agent temporarily search other regions (timesteps 100--375); (4) a combined \emph{tracking and re-acquisition} phase, where the NLL fluctuates rapidly as the agent iteratively loses and rediscovers the target (timesteps 375--500); (5) a transition back to \emph{exploration}, indicated by a gradual increase in NLL as the agent diverges to search frontier spaces (timesteps 500--750); (6) a final \emph{reacqisition} phase, where the NLL fluctuates repeatedly as the agent successfully re-acquires the target through active searching (timesteps 750--1000).
In contrast, the frontier-based expert (blue) continually expands its exploration frontier, demonstrating no explicit tracking or re-acquisition behavior, as evidenced by the brief drops in its curve. Meanwhile, the time-based expert (red) focuses heavily on a fixed tracking duration after detection, resulting in the highest minimum NLL among the three planners, while still maintaining a certain low level.

Fig.~\ref{fig:trajectories_map2} qualitatively compares the behaviors of the proposed and expert planners in the same episode as Fig.~\ref{fig:time-metrics}.
The frontier-based expert tends to focus on broad coverage, sweeping unexplored regions but not following already detected targets.
In contrast, MATT-Diff maintains target visibility through local tracking, occasionally performing short re-acquisition loops and then expanding its path toward unseen areas.
This results in a balanced behavior between exploration and tracking that corresponds to the temporal profile discussed above.

\begin{figure*}[t]
  \centering
  \includegraphics[
    width=0.99\linewidth
  ]{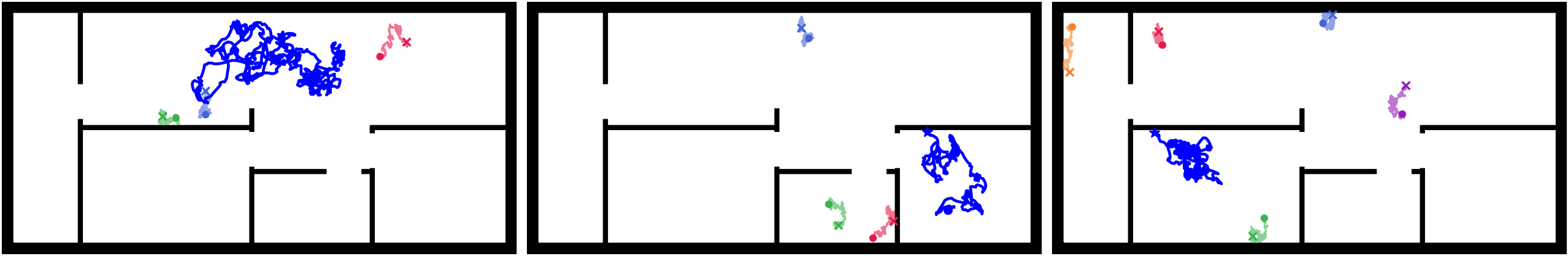}
  \caption{Trajectories followed by MATT-Diff across multiple episodes. The policy occasionally exhibits insufficient collision avoidance, causing the agent to get stuck in obstacle regions.}
  \label{fig:trajectories_collision}
\end{figure*}

We also observe certain failure modes depending on local geometric constraints. As illustrated in Fig.~\ref{fig:trajectories_collision}, MATT-Diff occasionally fails to maintain safety, resulting in collisions that trigger early termination. This effectively limits the agent's exposure to tracking and re-acquisition opportunities, whereas the expert planners consistently demonstrate robust collision avoidance.

\subsection{Limitations and Future Directions}

While MATT-Diff exhibits effective multi-modal behaviors for target tracking, several limitations remain.

\paragraph{Safety guarantee} Although the policy learns collision avoidance from expert demonstrations implicitly, it lacks explicit safety guarantees as observed in Fig.~\ref{fig:trajectories_collision}. While recent works have explored enforcing hard constraints on diffusion model outputs during inference \citep{romer2025diffusion, cheng2025safe}, integrating these mechanisms into the finetuning phase remains a challenge. An interesting question for future research is how to explicitly embed safeguards into the training loop such as policy learning through a constrained inference layer to ensure provably safe action generation.

\paragraph{Tuning the policy specifically for one map}
 Because diffusion policies inherently represent multi-modal action distributions, the specific mode executed at each timestep depends on stochastic sampling from the diffusion process. Fine-tuning the policy with reinforcement learning \citep{ren2025diffusion, wagenmaker2025steering} may help the agent learn exactly when to select each behavior mode, improving consistency and task performance. 

\paragraph{Training with other target estimators}
We remark that our policy is trained based on target estimation provided by a Kalman filter. As discussed in Section \ref{sec:problem}, this currently relies on a heuristic approach of abruptly removing the estimates of lost targets upon missing observations. More advanced target estimation methods that explicitly incorporate detection probabilities and missing observation models, such as the Probabilistic Hypothesis Density (PHD) filter \citep{vo2006gaussian}, can be used in the design. A key challenge for future work is to determine the network inputs and architecture capable of processing detection probabilities and data association probabilities provided by advanced probabilistic inference techniques.

\section{CONCLUSION} \label{sec:conclusion}

This paper proposed a novel network design and training methodology for active multimodal target tracking via a diffusion policy, named MATT-Diff. The exploration-exploitation dilemma faced in tracking targets with unknown number, states, and dynamics was addressed by MATT-Diff using demonstrations from three planners with distinct exploration and exploitation behaviors.
The network architecture of MATT-Diff applies CNN and vision-performer to the egocentric map and an attention mechanism to handle varying number of targets. The diffusion policy was trained to predict noise added to the expert action sequences given the observation sequences, and performed a denoising process starting from random noise to generate a multimodal action sequence. Our evaluation showed that MATT-Diff outperforms other learning-based approaches in an out-of-distribution environment and exhibits multimodal behavior derived from the expert planners.

\bibliography{bib/ref}

\end{document}